\documentclass[conference]{IEEEtran}
\IEEEoverridecommandlockouts
\usepackage{cite}
\usepackage{amsmath,amssymb,amsfonts}
\usepackage{algorithmic}
\usepackage{graphicx}
\usepackage{textcomp}
\usepackage{xcolor}
\usepackage{hyperref}
\usepackage{multirow}
\usepackage{cellspace}
\def\BibTeX{{\rm B\kern-.05em{\sc i\kern-.025em b}\kern-.08em
    T\kern-.1667em\lower.7ex\hbox{E}\kern-.125emX}}
\begin{document}

\title{LoRCoN-LO: Long-term Recurrent Convolutional Network-based
LiDAR Odometry\\
\thanks{All authors are with Seoul National University, South Korea, and D. Jung
is with Smart City Engineering major, and Integrated Major in
Smart City Global Convergence.}
}

\makeatletter
\newcommand{\linebreakand}{%
  \end{@IEEEauthorhalign}
  \hfill\mbox{}\par
  \mbox{}\hfill\begin{@IEEEauthorhalign}
}
\makeatother

\author{
\IEEEauthorblockN{
Donghwi Jung}
\IEEEauthorblockA{
\textit{Seoul National University}\\
Seoul, South Korea \\
donghwijung@snu.ac.kr}
\and
\IEEEauthorblockN{
Jae-Kyung Cho}
\IEEEauthorblockA{
\textit{Seoul National University}\\
Seoul, South Korea \\
jackyoung96@snu.ac.kr}
\and
\IEEEauthorblockN{
Younghwa Jung}
\IEEEauthorblockA{
\textit{Seoul National University}\\
Seoul, South Korea \\
xzxzmmnn@snu.ac.kr}
\linebreakand
\IEEEauthorblockN{
Soohyun Shin}
\IEEEauthorblockA{
\textit{Seoul National University}\\
Seoul, South Korea \\
soohyunshin@snu.ac.kr}
\and
\IEEEauthorblockN{
Seong-Woo Kim}
\IEEEauthorblockA{
\textit{Seoul National University}\\
Seoul, South Korea \\
snwoo@snu.ac.kr}
}

\maketitle

\begin{abstract}
We propose a deep learning-based LiDAR odometry estimation method called LoRCoN-LO that utilizes the long-term recurrent convolutional network (LRCN) structure. The LRCN layer is a structure that can process spatial and temporal information at once by using both CNN and LSTM layers. This feature is suitable for predicting continuous robot movements as it uses point clouds that contain spatial information. Therefore, we built a LoRCoN-LO model using the LRCN layer, and predicted the pose of the robot through this model. For performance verification, we conducted experiments exploiting a public dataset (KITTI). The results of the experiment show that LoRCoN-LO displays accurate odometry prediction in the dataset. The code is available at \url{https://github.com/donghwijung/LoRCoN-LO}.
\end{abstract}
\section{Introduction}
LiDAR is being widely used in autonomous driving \cite{jung2019curb, jung2021uncertainty, jung2022fast} based on its advantages such as a 360-degree field of view (FOV) and accurate distance measurement. In the case of LiDAR odometry estimation, existing methods have been mainly researched based on non-deep learning approaches.
Non-deep learning-based methods calculate the odometry between two sequences using ICP and NDT on a consecutive LiDAR point clouds \cite{hess2016real,koide2019portable,yokozuka2020litamin,park2018elastic,behley2018efficient}, or calculate the odometry after extracting specific features such as edge or planar features from point clouds \cite{zhang2014loam, shan2018lego, qin2020lins, shan2020lio}. In addition to the non-deep learning methods, recent research employ deep learning methods \cite{zheng2020lodonet, li2020dmlo, li2019net, nubert2021self, wang2019deeppco, cho2020unsupervised, wang2021pwclo}. 
A few of the studies have applied the deep learning model to specific parts such as key point matching \cite{zheng2020lodonet, li2020dmlo}, and some studies calculated odometry by using the point cloud as input data of the deep learning model in an end-to-end method \cite{li2019net, wang2021pwclo, nubert2021self, wang2019deeppco, cho2020unsupervised}. In the case of point cloud data, it was used directly in 3D \cite{wang2021pwclo} or after being projected in 2D \cite{zheng2020lodonet, li2020dmlo, li2019net, nubert2021self, wang2019deeppco, cho2020unsupervised}.\\
\indent Among the non-deep learning based methods, the ICP (or NDT)-based has high accuracy in that it uses almost all points. However, there is a disadvantage in that it is hard to guarantee real-time performance because the operation speed is relatively slow. In the case of feature-based, the main purpose is to predict changes between sequences through feature extraction, thus feature dependence is high. For example, for edge or planar features used in LOAMs \cite{zhang2014loam, shan2018lego, qin2020lins, shan2020lio}, it is difficult to extract features when there are few structures such as buildings or columns, such as off-road. These characteristics make the algorithm difficult to apply. In addition, the non-deep learning-based method requires a lot of computation, hence the computational cost is high compared to the deep learning-based method. On the other hand, in the case of existing deep learning-based methods, if only model training is properly performed, high-accuracy results can be obtained at high speed by utilizing GPU. At this time, the previous deep learning-based methods mainly utilize CNN and fully connected layers to calculate only the odometry between two consecutive sequences.\\
\begin{figure*}[t]
    \centering
    \framebox{\parbox{0.9\textwidth}{\includegraphics[width=0.9\textwidth]{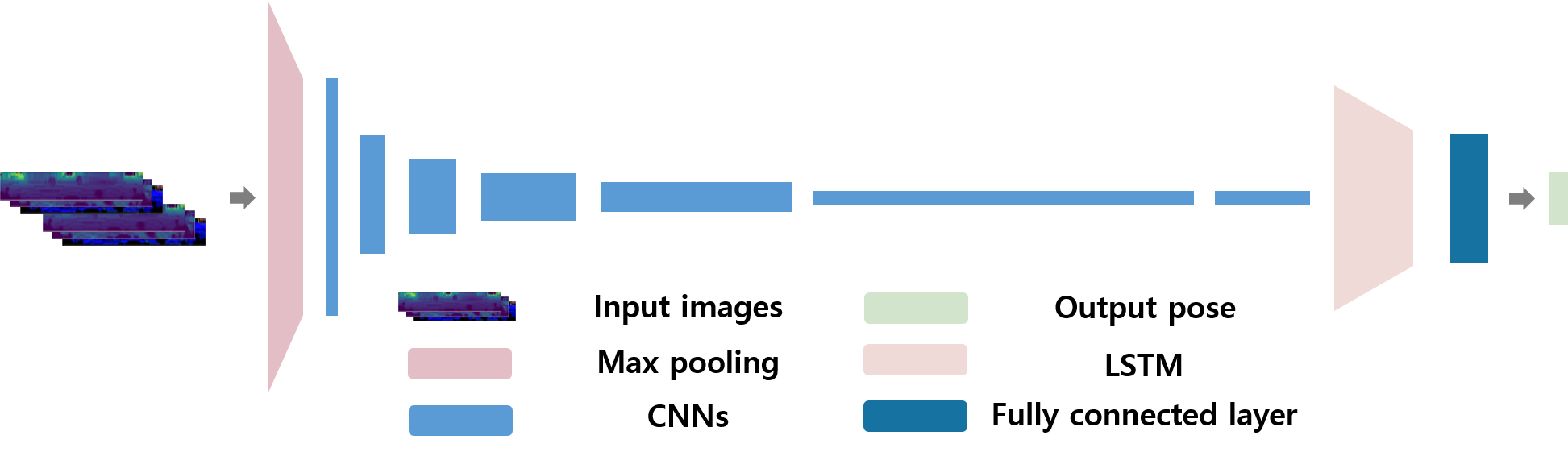}}}
    \caption{System architecture of LoRCoN-LO. Input is 2D projected point cloud image, and the output is a 6-DOF robot pose.}
    \label{fig:architecture}
\end{figure*}
\indent However, the movement of a real robot is continuous, thus the current motion is related to the previous motions. Therefore, when predicting the current movement, it is helpful to improve the performance by utilizing the information of the past movements. For this reason, in this paper, we propose LoRCoN-LO, a deep learning-based LiDAR odometry estimation algorithm using an LRCN proposed in the paper \cite{donahue2015long}. The LRCN layer is a structure that can process spatial and temporal information at once by using both CNN and LSTM structures. This feature is suitable for predicting continuous robot movement using a point cloud containing spatial information. Therefore, we constructed a LoRCoN-LO model using the LRCN layer, and predicted the robot pose through this model. In this paper, we used the KITTI dataset to test the model. As a result, LoRCoN-LO represents accurate odometry prediction performance.
\begin{figure*}[t]
    \centering
    \framebox{\parbox{\textwidth}{\includegraphics[width=\textwidth]{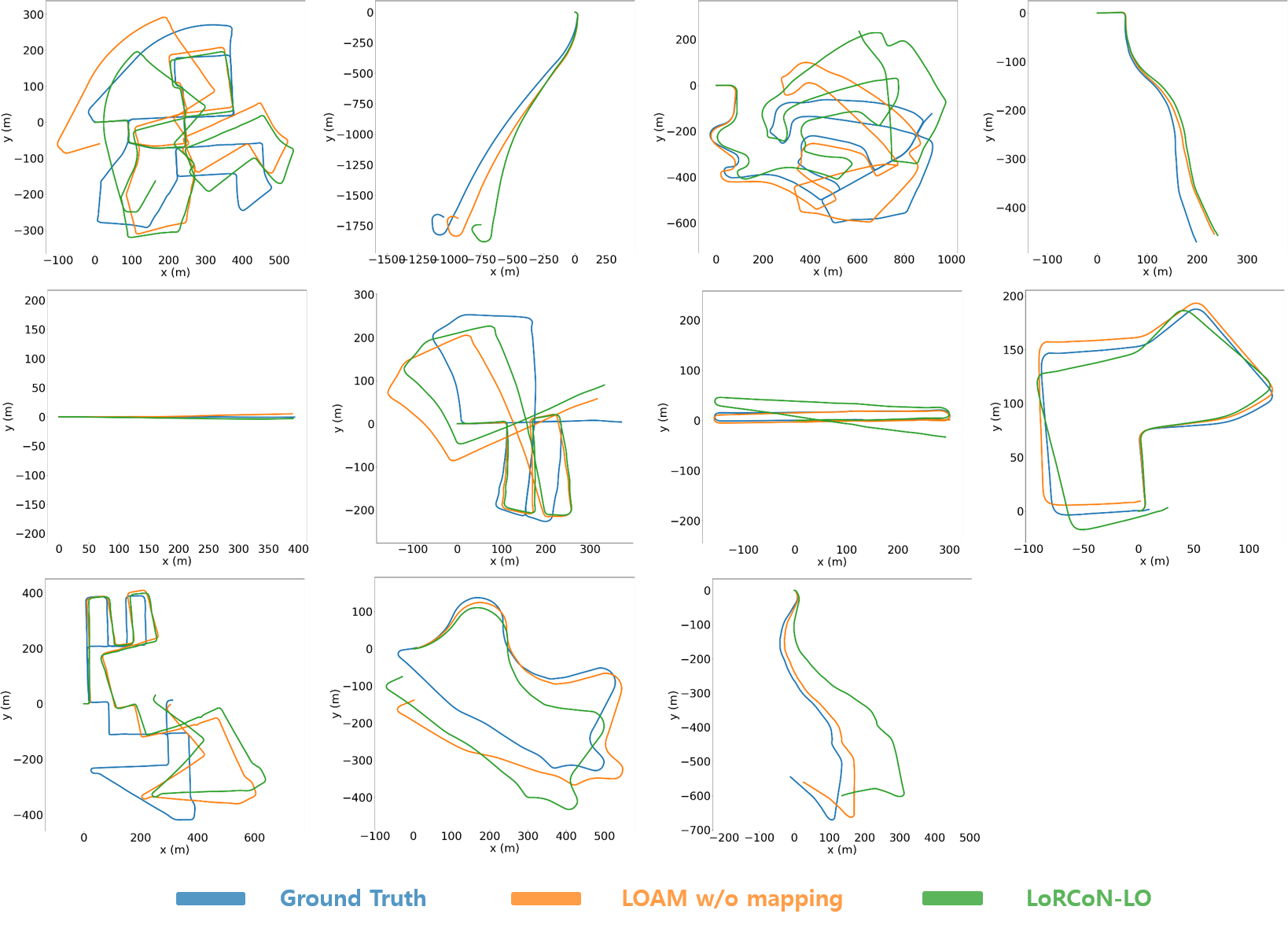}}}
    \caption{Visualized results of odometry estimation with KITTI dataset.}
    \label{fig:kitti_path_result}
\end{figure*}
\section{Methods}
\subsection{Data Pre-Processing}\label{data_pre_processing}
The 3D point cloud is projected as a 2D image to be used as input data of LoRCoN-LO. This projection corresponds to the conversion of coordinates from $x$, $y$, $z$ to $u$, $v$. In the process of projection, the vertical resolution of LiDAR is used for the width, and the horizontal resolution of LiDAR is used for the height. For points with the same $x$ and $y$ values, the closest point from the LiDAR is selected as the corresponding pixel.
\begin{gather}
    (x,y,z)\rightarrow(u,v),\\
    \label{eqn:distance}
    d=\sqrt{x^2+y^2+z^2},\\
    \label{eqn:2d_projection}
    \begin{pmatrix}u\\v\end{pmatrix}=\begin{pmatrix}\frac{1}{2}\cdot\left[1-\pi^{-1}\cdot\arctan{\left(y,x\right)}\right]\cdot w\\\left[1-\left(\arcsin{\left(z\cdot d^{-1}\right)}+f_{up}\right)\cdot f^{-1}\right]\cdot h\end{pmatrix},\\
    \vec{n}=\frac{\vec{a}\times\vec{b}}{|\vec{a}|\cdot|\vec{b}|\cdot\sin{\theta}},
\end{gather}
where $(x,y,z)$ and $u, v$ denote 3D point cloud and 2D projected image coordinates. Moreover, $x, y, z$ in Eq. \eqref{eqn:distance} correspond to the $x, y, z$ values for the point in a point cloud. Therefore, $d$ indicates the distance from the LiDAR. In addition, $w$ and $h$ are width and height of the 2D image. $f_{up}$ and $f_{down}$ denote minimum and maximum of vertical filed-of-view of LiDAR sensor. Furthermore, $f$ indicates sum of $f_{up}$ and $f_{down}$. The 3D point cloud is projected as a 2D image according to Eq \eqref{eqn:2d_projection}. For Eq \eqref{eqn:2d_projection}, we referred to the spherical projection described in the paper \cite{behley2018efficient}. Additionally, $\vec{a}$ and $\vec{b}$ are vectors that connect the center pixel and the nearest neighbor pixels. $\theta$ denote the angle between two vectors $\vec{a}$ and $\vec{b}$. Furthermore, $\vec{n}$ is the unit vector that is perpendicular to the plane containing two vectors $\vec{a}$ and $\vec{b}$.\\
\indent Three types of 2D images (\emph{depth}, \emph{intensity}, and \emph{normal}) are used as an input data. For depth, it corresponds to the distance $d$. In the case of intensity, the intensity value of each point obtained by LiDAR is used. For the normal, considering the real-time of the calculation, we find the nearest neighbors (pixels) in pixel unit in the 2D image and compute the normal vector $\vec{n}$ by performing cross product using the $x$, $y$, $z$ values between the pixels and the center pixel. Finally, to represent in the form of an image, the $x$, $y$, and $z$ values of the normal vector are matched to the $r$, $g$, and $b$ values of the image, respectively.
\begin{table*}[t]
\caption{Results of odometry estimation with KITTI dataset.}
\label{table:kitti_result}
\begin{center}
\resizebox{0.9\textwidth}{!}{
    \begin{tabular}{Sc Sc Sc Sc Sc Sc Sc Sc Sc Sc Sc Sc Sc Sc Sc Sc Sc Sc Sc Sc Sc Sc}\hline
    &&00&01&02&03&04&05&06&07&08&09&10\\\hline
    \multirow{2}{*}{LOAM \cite{zhang2014loam}}&$t_{rel}$&5.36&\textbf{5.45}&\textbf{4.10}&4.68&1.82&3.47&5.68&4.64&\textbf{4.04}&\textbf{9.88}&9.68\\
    &$r_{rel}$&2.23&\textbf{1.91}&\textbf{1.79}&2.53&1.46&1.60&2.14&2.45&\textbf{1.75}&\textbf{3.78}&\textbf{4.40}\\\hline
    \multirow{2}{*}{Our method}&$t_{rel}$&\textbf{4.71}&5.96&4.74&\textbf{3.44}&\textbf{1.66}&\textbf{2.77}&\textbf{3.74}&\textbf{3.14}&5.07&11.08&\textbf{7.18}\\
    &$r_{rel}$&\textbf{2.16}&2.12&1.83&\textbf{1.63}&\textbf{1.42}&\textbf{1.43}&\textbf{1.43}&\textbf{2.11}&2.15&4.13&5.16\\\hline
    \end{tabular}
}
\end{center}
\end{table*}
\subsection{Architecture}
The overall architecture of the model is shown in Fig. \ref{fig:architecture}. A detailed description is as follows:\\
\subsubsection{Input Data}\label{input_data}
Input data consists of a total of 10 channels by connecting 2 consecutive data of 5 channels. In one of these sequences, the three types of input images form a total of 5 channels (\emph{Depth} 1 channel, \emph{Intensity} 1 channel, \emph{Normal} 3 channels). At this time, the difference in width and height of the image projected in 2D occurs due to the difference in resolution between axes of LiDAR. Width is relatively longer than height. Therefore, if the image is used as it is, it will have almost zero height after several CNN layers. To prevent this, first we apply max pooling to reduce the width in half.\\
\subsubsection{CNN Layers}
To embed the image data, a 6-layer CNN is used. At this time, considering that the 2D panoramic image is a 360-degree point cloud projection, we apply circular padding between each layer. Moreover, for the purpose of regularization, batch normalization is added between each layer. In addition, as we mentioned in Sec. \ref{input_data}, width is relatively longer than height. Therefore, a stride that is bigger than 1 for the horizontal axis is utilized except for the last layer.\\
\subsubsection{LSTM Layer}
The data passed through the CNN layers goes into the LSTM layer. The LSTM layer is set up as a deep bi-directional LSTM consisting of four layers to increase the prediction accuracy. Moreover, for avoiding overfitting, dropout is used after the LSTM layer.\\
\subsubsection{Fully Connected Layer}\label{fully_connected_layer}
A fully connected layer is connected to predict the robot pose with 6-degree of freedom (DOF) (x,y,z for translation, roll, pitch, yaw for rotation). When training the model, data from all sequences of the LSTM layer are used to predict the robot pose and reflected it in learning. However, in the case of inference, only the data from the last sequence is used to predict the robot pose.\\
\section{Experiments}
In this paper, the KITTI dataset was used to test the algorithm performance, and the results were compared with the baseline. Moreover, for the baseline, LOAM was chosen to compare the performance of LoRCoN-LO, as LOAM is a feature-based LiDAR SLAM which until recently performed well on the KITTI odometery benchmark \cite{geiger2012we}. In addition, because the code is publicly available, it is currently being used as a baseline in many studies. For metrics, the average translation and rotation errors in the Table. \ref{table:kitti_result} were calculated to compare the odometry estimation performances in the KITTI dataset. These errors were computed using the calculation method suggested by the KITTI benchmark \cite{geiger2012we}. It was measured as a root mean squared error (RMSE), and the overall average was obtained after calculating each error by increasing the path interval from 100m to 800m in 100m increments. Moreover, it displays the error in \% for translation and in degrees per 100 m for rotation. On the other hand, unlike Table. \ref{table:kitti_result}, RMSE of instantaneous odometry was calculated and compared without accumulation of odometry. We applied this additional metric to check how large the error of cumulative odometry can be compared to single odometry error.\\
\indent All code implementations were done using \emph{PyTorch}. Moreover, for the 2D panoramic images described in Sec. \ref{data_pre_processing}, 64 was used for the height corresponding to the number of LiDAR channels, and 900 was applied to the width with a resolution of 0.4 degrees per pixel. In addition, 4 was selected as the LSTM sequence length, and a probability of 0.5 was applied to the dropout placed after the LSTM layer. Furthermore, we set the batch size to 32, used the Adagrad optimizer, and set the learning rate to 0.0005 for training the model. Additionally, the loss function was calculated as the mean squared error (MSE). Because, the rotation error has a greater effect on the odometry prediction accuracy than the translation error, the rotation error is multiplied by a weight of 100. For training, 400 epochs were performed on the KITTI dataset. Lastly, to compare equally with LOAM, training and testing were conducted by changing all poses to LiDAR coordinates.\\
\indent Among the sequences of KITTI dataset, sequences 09 and 10 were used as test data, and the remainders from 00 to 08 were used as training data. As shown in Table \ref{table:kitti_result}, the odometry prediction results show high accuracy in odometry estimation. This is confirmed not only in the sequences 00-08 used as the training data, but also in the sequences 09, 10 used as the test data. Moreover, as shown in Fig. \ref{fig:kitti_path_result}, LoRCoN-LO represents better performance not only in sequences composed of almost straight paths like 04, but also in sequences containing many rotations, such as 00 or 05. Therefore, we can conclude that LoRCoN-LO provides accurate odometry in both translations and rotations.
\section{Conclusion}
In this paper, we propose a LoRCoN-LO which is a deep learning-based LiDAR odometry estimation method using LRCN structure. The LRCN layer is a structure that can process spatial and temporal information at once by using both CNN and LSTM structures. This feature is suitable for predicting continuous robot movements using a point cloud containing spatial information. To evaluate the performance of the LoRCoN-LO model using the LRCN layer, the performance of the model was tested in KITTI dataset. As a result, we confirmed that the model can predict LiDAR odometry with good accuracy.
\section*{Acknowledgement}
This work was supported by Korea Institute for Advancement of Technology(KIAT) grant funded by the Korea Government(MOTIE) (P0020536, HRD Program for Industrial Innovation), by the National Research Foundation of Korea through the Ministry of Science and ICT under Grant 2021R1A2C1093957, by Korean Ministry of Land, Infrastructure, and Transport as Innovative Talent Education Program for Smart City, by the Daewoo Shipbuilding \& Marine Engineering through the Future Ocean Cluster, and by the Institute of Engineering Research at Seoul National University.

\bibliographystyle{IEEEtran}
\bibliography{IEEEabrv, main}
\end{document}